\documentclass[11pt,a4paper]{article}
\usepackage[hyperref]{emnlp-ijcnlp-2019}
\usepackage{times}
\usepackage{latexsym}

\usepackage{url}
\usepackage{graphicx}
\usepackage{float}
\usepackage{subfigure}
\usepackage{amsmath}
\usepackage{amsfonts}
\usepackage{multirow}
\usepackage{enumerate}

\aclfinalcopy 

\title{Modeling Inter-Aspect Dependencies with a Non-temporal \\Mechanism for Aspect-Based Sentiment Analysis}
\author{
  Yunlong Liang\textsuperscript{1}\thanks{ \ \ Work was done when Yunlong Liang was an intern at Pattern Recognition Center, WeChat AI, Tencent Inc, China.}  , 
  Fandong Meng\textsuperscript{2}, 
  Jinchao Zhang\textsuperscript{2}, 
  {Yufeng Chen}\textsuperscript{1} \\
  \textbf{Jinan Xu}\textsuperscript{1}\thanks{ \ \ Jinan Xu is the corresponding author.} and
  \textbf{Jie Zhou}\textsuperscript{2}\\
  \textsuperscript{1}Beijing Jiaotong University, China \\
  \textsuperscript{2}Pattern Recognition Center, WeChat AI, Tencent Inc, China \\
  \texttt{\{yunlonliang,chenyf,jaxu\}@bjtu.edu.cn} \\
  \texttt{\{fandongmeng,dayerzhang,withtomzhou\}@tencent.com} \\
}
\date{}

\begin{document}
\maketitle
\begin{abstract}
  For multiple aspects scenario of aspect-based sentiment analysis (ABSA), existing approaches typically ignore inter-aspect relations or rely on temporal dependencies to process aspect-aware representations of all aspects in a sentence. Although multiple aspects of a sentence appear in a non-adjacent sequential order, they are not in a strict temporal relationship as natural language sequence, thus the aspect-aware sentence representations should not be treated as temporal dependency processing. In this paper, we propose a novel non-temporal mechanism to enhance the ABSA task through modeling inter-aspect dependencies. 
  Furthermore, we focus on the well-known class imbalance issue on the ABSA task and address it by down-weighting the loss assigned to well-classified instances. Experiments on two distinct domains of SemEval 2014 task 4 demonstrate the effectiveness of our proposed approach. 
\end{abstract}

\section{Introduction}

Aspect-based sentiment analysis (ABSA) is a fine-grained task of sentiment classification, which aims to detect the sentiment polarity towards a given target aspect. There may be single or multiple aspects in a sentence. 
For the sentence containing multiple aspects, existing models~\citep{Wang:16,Ma:17,Yi:17,he-etal-2018-effective,Huang:18:PCNN:b,Fan:18,weixueGCAE:18,li2018transformation,WangS:18,DBLP:journals/corr/abs-1811-10999,liang-etal-2019-novel,liang-etal-2021-iterative-multi,liang2020dependency,DBLP:journals/corr/abs-1904-02232}
typically generate the aspect-aware sentence representation for classification without considering the influence caused by the surrounding aspect. 

However, some work~\citep{2018-modeling-inter-aspect,Majumder:18} shows that aspect-aware sentence representations of the neighboring aspect is beneficial for sentiment predictions of the target aspect. 
For instance, {\em ``\underline{Service} was \textbf{good} and so was the \underline{atmosphere}.''}, the polarity of the aspect {\em ``atmosphere''} is influenced by the sentiment of the aspect {\em ``service''}. 
~\citet{2018-modeling-inter-aspect} first find such phenomena and utilize temporal dependency to process the aspect-aware sentence representation. 
More recently, based on the temporal dependency processing, ~\citet{Majumder:18} extend ~\citet{2018-modeling-inter-aspect}'s work with the attention mechanism~\cite{D15-1166} and memory network~\cite{Jason:WestonCB14}. Although those aspects of a sentence appear in a non-adjacent sequential order, they are not in a strict temporal relationship as natural language sequence. 
We thus argue that the aspect-aware sentence representation should not be treated as temporal dependency processing. 

In addition, we also focus on the class imbalance issue. For instance, {\em ``\underline{Desserts} include \underline{flan} and \underline{sopaipillas.}''}, polarities of all aspects (underlined) are ``neutral''. But their polarities are easy to be wrongly induced to ``positive'' by existing models. This may be caused by the class imbalance issue. 
It is well-known that the datasets of SemEval 2014 task 4~\citep{Pontiki:14} have the class imbalance issue in the training example (e.g., $class_{positive}:class_{neutral}=3.4:1$ on restaurant domain), which makes the classifier tend to predict ``positive''. 

To address those issues mentioned above, in this paper, we propose a non-temporal mechanism to model inter-aspect dependencies. 
Firstly, we independently generate all aspect-aware sentence representations. 
Then, we utilize the non-temporal mechanism to control how much the surrounding aspect-related information flow into the target-specific representation. Moreover, we introduce the focal loss~\citep{DBLP:journals/corr/abs-1708-02002}, which was first proposed in computer vision, to address the class imbalance issue by down-weighting the loss assigned to well-classified instances.

We evaluate the effectiveness of our approach on two distinct domains of SemEval 2014 task 4. Experiment results suggest that the non-temporal mechanism can effectively integrate the neighboring aspect-related information, conducting more accurate predictions. Furthermore, the focal loss can substantially mitigate the class imbalance issue and further improve the performance. We also provide empirical analysis to reveal the advantages of our proposed approach. Our contributions can be summarized as follows: 
\begin{itemize}
\item We propose a novel non-temporal mechanism to enhance the ABSA task through modeling inter-aspect dependencies, which can effectively integrate the neighboring aspect-related information.
\item To our best knowledge, we are the first that introduce focal loss to address the class imbalance issue for the ABSA task.
\item Our approach has shown its excellent performances on two distinct domains. 
\end{itemize}

\section{Approach}
\subsection{Problem Definition}
Giving a sentence \emph{S} = \{${w}_{1},{w}_{2},...,{w}_{n}$\}, where ${w}_{i}$ is the $i$th word and ${n}$ is the sentence length, it may have multiple aspects ${A}$ = \{${a}^{t},{a}_{1},{a}_{2},...,{a}_{m}$\}, where $|m+1|$ is the number of all aspects and ${a}_{i}$ (${a}^{t}$) is a subsequence with $k$ words of the sentence, i.e., ${a}_{i}$ = \{${w}_{q},{w}_{q+1},...,{w}_{q+k-1}$\}, $1 \leq q \leq n~$ and $~1 \leq k \leq n-q+1$.
The goal of the ABSA task is to predict the polarity 
for the target aspect.
\subsection{Aspect-Aware Sentence Representation}
\begin{figure*}[!htb]
\centering
  \includegraphics[width = 1.0\textwidth]{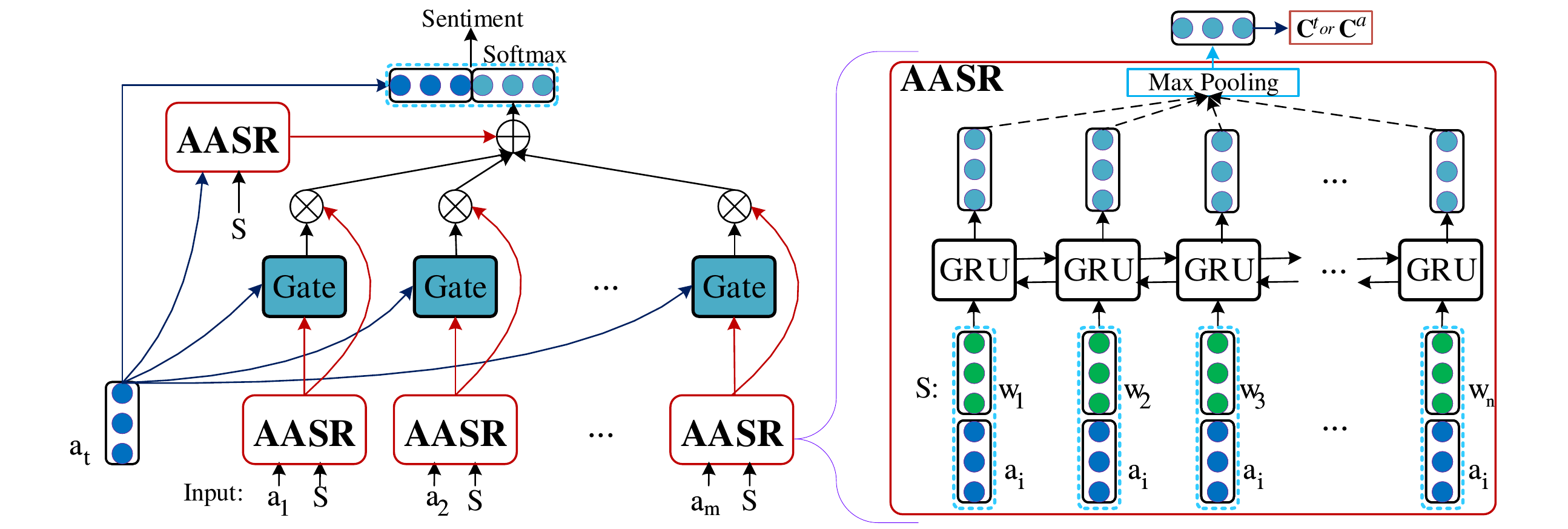}
\caption{Overview of our architecture. AASR~\citep{Majumder:18} represents \emph{Aspect-Aware Sentence Representation}.}
\label{fig:gate}
\end{figure*}

Our architecture is shown in Figure~\ref{fig:gate}, where we employ gated recurrent unit (GRU)~\citep{DBLP:journals/corr/ChungGCB14} as the encoder. We concatenate the aspect representation with every word embedding as the input of the encoder. In order to obtain the global context information of the sentence towards the given aspect, we utilize the GRU, which is described as follows:
\begin{align}
  \label{eq:gru_h}
  \mathbf{h}_{t} &= (1 - \mathbf{z}_{t}) \odot \mathbf{h}_{t-1} + \mathbf{z}_{t} \odot \widetilde{\mathbf{h}}_{t} \\
    \label{eq:gru_h_}
  \widetilde{\mathbf{h}}_{t} &= \text{tanh}(\mathbf{W}_{x}\mathbf{x}_{t} + \mathbf{r}_{t} \odot (\mathbf{W}_{h}\mathbf{h}_{t-1}))
\end{align}
where $\mathbf{x}_{t}$ denotes the input embedding of time step $t$; the update gate $\mathbf{z}_{t}$ and the reset gate $\mathbf{r}_{t}$ are computed as:
\begin{align}
  \label{eq:gru_r}
  \mathbf{r}_{t} &= \sigma(\mathbf{W}_{xr}\mathbf{x}_{t} + \mathbf{W}_{hr}\mathbf{h}_{t-1})\\
  \label{eq:gru_z}
  \mathbf{z}_{t} &= \sigma(\mathbf{W}_{xz}\mathbf{x}_{t} + \mathbf{W}_{hz}\mathbf{h}_{t-1})
\end{align}
Furthermore, a forward GRU is applied to generate the hidden vector \{$\overrightarrow{\mathbf{h}_{1}},\overrightarrow{\mathbf{h}_{2}},...,\overrightarrow{\mathbf{h}_{n}}$\} and a backward GRU is applied to obtain the hidden vector \{$\overleftarrow{\mathbf{h}_{1}},\overleftarrow{\mathbf{h}_{2}},...,\overleftarrow{\mathbf{h}_{n}}$\}. Subsequently, we get the final representation by concatenating two vectors: $\mathbf{h}_{i}$ = [$\overrightarrow{\mathbf{h}_{i}},\overleftarrow{\mathbf{h}_{i}}$] and leverage max pooling to obtain the aspect-aware sentence representation for the aspect.
\subsection{MIAD: Modeling Inter-Aspect Dependencies}
After the AASR module (on the right of Figure~\ref{fig:gate}) , we can obtain all aspect-aware sentence representations \{$\mathbf{C}^t,\mathbf{C}_{1}^a,\mathbf{C}_{2}^a,...,\mathbf{C}_{m}^a$\} where $\mathbf{C}^t$ is the \textbf{t}arget aspect-aware sentence representation and $\mathbf{C}_{i}^a$ is its surrounding \textbf{a}spect-specific sentence representation in a sentence. In order to integrate the neighboring aspect related information with a non-temporal manner, we design gates for every surrounding aspect-aware sentence representation. Each gate is designed as follows:
\begin{align} 
  \label{eq:aspect_gate}
  \mathbf{\hat{g}}^{{a}}_{i} = \sigma(\mathbf{W}_{a}\mathbf{a}^{t} + \mathbf{W}_{cg}\mathbf{C}^{a}_{i})
\end{align}
where $\mathbf{a}^{t}$ denotes \textbf{t}arget aspect embedding. Then ${\mathbf{\hat{g}}}^{{a}}_{i}$ are normalized as:
\begin{equation} 
  \label{eq:aspect_gate_}
  \mathbf{g}^{{a}}_{0},\mathbf{g}^{{a}}_{1},...,\mathbf{g}^{{a}}_{m} = \text{softmax}(\mathbf{\hat{g}}^{{a}}_{0}, \mathbf{\hat{g}}^{{a}}_{1},...,\mathbf{\hat{g}}^{{a}}_{m})
\end{equation}

Then, we utilize those well-designed gates to control how much the neighboring aspect related information is included in the final target-specific representation as follows:
\begin{equation}
  \label{eq:H_t1} 
  \mathbf{C}^{f} = \mathbf{C}^{t} + \sum_{i=0}^{m}( \mathbf{g}^{{a}}_{i} \odot \mathbf{C}^{{a}}_{i})
\end{equation}
where $\odot$ denotes element-wise product. 

Clearly, $\mathbf{C}^{f}$ has selectively assimilated the neighboring aspect related information in a non-temporal manner. Subsequently, the final fully-connected layer with softmax function takes the target-specific representation $\mathbf{C}^{f}$ as input to predict the sentiment polarity ${p}_{i}$ for the target aspect.

\subsection{Objective Function}
In order to address the class imbalance issue, we utilize a new loss function, namely focal loss~\citep{DBLP:journals/corr/abs-1708-02002}, which is adapted from standard cross entropy loss. The adapted focal loss is written as:
\begin{equation}
    \label{eq:L_fl}
  \mathcal L_{FL} = - \sum_{i=0}^{L}{y}_{i} (1 - {p}_{i})^\gamma \log({p}_{i}) 
\end{equation}
where $L$ is the number of sentiment labels, ${y}_{i}$ and ${p}_{i}$ is the ground truth and the estimated probability for the sentiment label, respectively. The focusing parameter $\gamma$ smoothly adjusts the rate at which easy instances are down-weighted. If an instance is misclassified and ${p}_{i}$ is small, the modulating factor will be near 1 and the loss is unaffected. when ${p}_{i} \rightarrow{1}$, the factor will be near to 0 and thus the loss of well-classified instances will be down-weighted.

When generating neighboring aspect-aware sentence representations (\{$\mathbf{C}_{1}^a,\mathbf{C}_{2}^a,...,\mathbf{C}_{m}^a$\}), we also design loss function as follows:
\begin{equation}
  \begin{split}
  \label{eq:neighboring_aspect_loss}
  \mathcal L_{NA} =  - \frac{1}{m}\sum_{j=0}^{m}\sum_{i=0}^{L}{y}_{i}^{j}(1 - {p}_{i}^{j})^\gamma\log({p}_{i}^{j})
  \end{split}
\end{equation}
where $m$ is the number of neighboring aspects in a sentence,  ${y}_{i}^{j}$ and ${p}_{i}^{j}$ is the ground truth and the estimated probability of the sentiment label for $i$th aspect, other symbols are the same as those in $\mathcal L_{FL}$.

\begin{table}[h]
\centering
\setlength{\tabcolsep}{1.1mm}{
\begin{tabular}{l|l|ll|ll|ll}
\hline
\multirow{2}{*}{} & \multicolumn{1}{c|}{} & \multicolumn{2}{c|}{\textbf {Positive}} & \multicolumn{2}{c|}{\textbf {Negative}} & \multicolumn{2}{c}{\textbf {Neutral}} \\
\cline{3-8}
                          &                  & SA         & MA        & SA        & MA         & SA        & MA   \\ \hline
\multirow{2}{*}{$\textbf{Laptop}$}& Train    & 349        & 638       & 442       & 424        & 126       & 334     \\
                                & Test       & 137        & 204        & 69       & 59         & 53        & 116    \\ \hline
\multirow{2}{*}{$\textbf{Rest.}$} & Train    & 609        & 1,555     & 226       & 579        & 173       & 460     \\
                                  & Test     & 182        & 546        & 62       & 134         & 41       & 155       \\ \hline

\end{tabular}}
\caption{Distribution of the dataset by class labels and single aspect/multiple aspect in SemEval 2014. `Rest.': Restaurant, `SA': Single Aspect, `MA': Multi-Aspect.}
\label{tbl:dataset}
\end{table}
Thus, our final objective function is written as:
\begin{equation}
    \label{eq:J_loss}
  \boldsymbol{J} = min (\mathcal L_{FL} + {\lambda}\mathcal L_{NA})
\end{equation}
where ${\lambda}$ is the weight of loss $\mathcal L_{NA}$.

\section{Experiments}

\subsection{Setup}
\paragraph{Datasets.} We perform experiments on the datasets of SemEval 2014 task 4, which contains two distinct domains: restaurant and laptop. Table~\ref{tbl:dataset} shows the distribution of the datasets.

\begin{table*}[t!]
\begin{center}
\setlength{\tabcolsep}{0.4mm}{
\begin{tabular}{l|l|cccccc|cccccc}
\hline
\multirow{2}{*}{}& & \multicolumn{6}{c|}{$\textbf{Laptop}$} & \multicolumn{6}{c}{$\textbf{Restaurant}$}  \\
\cline{3-14}
         &  & Total        & SA        & MA         & Neu          & Neg       & Pos     & Total        & SA        & MA        & Neu          & Neg       & Pos   \\ \hline
\multirow{3}{*} {\textbf{Baselines}} &\textbf{MIA}~\citep{2018-modeling-inter-aspect} & 72.5        & -            & -             & - & -   & -         & 79.0              & -   & -             & - & -   & -   \\
&\textbf{IAN}~\citep{Ma:17}* &72.1       & 72.5            & 71.6             & - & -   & -   & 78.6              & 75.4   & 77.7             & - & -   & -          \\
&\textbf{IARM}~\citep{Majumder:18} & 73.8 & 73.4   & 74.1  & - & -   & - & 80.0   & 78.6  & 80.48  & - & -  & -\\ \hline \hline
\multirow{5}{*}{\textbf{Ours}}
 &\textbf{GRU} & 71.6   & 71.8    & 71.5    & 46.2  & 64.1   & 87.1         & 79.1   & 79.3    & 79.0   & 36.2  & 61.7   & \textbf{95.3}       \\
 &\textbf{GRU+TM} & 72.3   & 71.8    & 72.6    & 50.3  & 57.0   & 88.9         & 79.8   & 79.6    & 79.9   & 33.2  & 71.9   & 94.5       \\
 &\textbf{GRU+NoTM} & 73.4   & 71.1    & 74.9    & 44.4    & 67.2  & \textbf{90.0}         & 80.5   & 79.3   & 80.8   & 39.8  & \textbf{75.0}   & 92.9     \\
 &\textbf{GRU+FL}  & 73.1   & 72.6    & 73.4   & \textbf{57.4}  & {68.0}   & 82.7           & 80.4   & 80.0  & 80.5  & 42.3  & 67.9   & 94.0     \\
 &\textbf{GRU+NoTM+FL (MIAD)}& \textbf{75.3} & \textbf{73.8} & \textbf{76.3}  & 55.7  & \textbf{70.3}   & 86.8 & \textbf{81.0}  & \textbf{80.7}  & \textbf{81.1}  & \textbf{49.0} & 63.3   & 94.4
\\ \hline
\end{tabular}}
\end{center}
\caption{The accuracy of multiple scenarios. `*' denotes the result is retrieved from~\citet{Majumder:18}. `GRU' denotes without considering the neighboring aspects. `TM' and `NoTM' indicates considering neighboring aspects with temporal dependency processing and non-temporal mechanism, respectively. `FL' represents focal loss. `Neu': Neutral, `Neg': Negative, `Pos': Positive.}
\label{tbl:Ablation_result}
\end{table*}
\paragraph{Training Details.}
300d Glove is adopted to initialize word embeddings~\citep{glove:14}. 
For optimization, we use the Adam optimizer~\citep{Adam:14} with initial learning rate 0.01. 
Focusing parameter $\gamma$ is set to 2.0. Weight of the loss ${\lambda}$ in Eq.~\ref{eq:J_loss} is set to 0.4, 0.2 for restaurant and laptop domain, respectively. Evaluation metrics are accuracy.

\subsection{Results and Analysis}
To comprehensively compare our method with baselines, we conduct three scenarios experiments. We name our architecture as MIAD. 

\paragraph{Domain-Wise Comparison.}

On both domains in Table~\ref{tbl:Ablation_result} (`Total' part), our method consistently outperforms all baseline methods. IAN ignores the surrounding aspect. MIA firstly models the inter-aspect relation with temporal dependency processing and IARM extends MIA's work with attention and memory network. MIAD surpasses IARM by 1.5\% on restaurant domain and by 1.0\% on laptop domain. This demonstrates that the combination of the non-temporal mechanism and the focal loss has a significantly positive effect on prediction process. 

\paragraph{Single Aspect and Multi-Aspect Scenarios Evaluation.}
In `SA' and `MA' parts of Table~\ref{tbl:Ablation_result}, our MIAD beats the IAN and IARM models. And our method ``GRU+NoTM'' obtains slight gains against IARM, especially in `MA' part. The reason may be that IARM is armed with multiple attentions and memory networks (we only apply GRU). 
Based on our settings, we implement the idea of temporal dependency processing method (``GRU+TM''). ``GRU+TM'' and ``GRU+NoTM'' methods highly surpass the baseline ``GRU'' in `MA' part, which shows that the neighboring aspect is beneficial for target-aspect sentiment prediction and this is consistent with the previous work~\citep{2018-modeling-inter-aspect,Majumder:18}. ``GRU+NoTM'' also gives significantly
better accuracies compared with ``GRU+TM'' in `MA' part.  Results suggest that the non-adjacent sequential order should not be treated as temporal dependencies processing indeed, and our non-temporal mechanism is highly competent to process this. 
\paragraph{Class-Domain Evaluation.}
It is evident that the class imbalance issue emerges in Table~\ref{tbl:dataset}. Accordingly, the result in Table~\ref{tbl:Ablation_result} (`Neu', `Neg' and `Pos' parts) is consistent with it. 
This suggests that the classifier tends to classify the sentiment polarity to be ``positive'', and the focal loss that can significantly mitigate this issue (``GRU+FL'' vs. ``GRU'').
\subsection{Case Study}
We now give some real examples to reveal the capability of our approach. 
The sentence {\em``Probably my \textbf{worst} \underline{dining experience} in new york, and I'm a \underline{former} waiter so I know what I'm talking about.''} with aspect {\em``former''} and ``neutral'' sentiment, fails to be correctly classified by ``GRU+TM''. 
Since, ``GRU+TM'' transfers the ``negative'' sentiment of the former aspect 
due to the temporal dependencies processing. On the other hand, ``GRU+NoTM'' succeeds in this case with the non-temporal mechanism, which can effectively integrate neighboring aspects related information and thus lead to correct prediction. 

Another case {\em``\textbf{Great} \underline{beer selection} too, something like 50 \underline{beers}.''} contains two aspects 
with corresponding sentiments: ``positive'' and ``neutral''.
Here, ``GRU'' fails to make correct prediction for the aspect {\em``beers''} due to the influence of the former aspect or the class imbalance issue, while MIAD makes correct final classification. 
This benefits from the non-temporal mechanism that can block irrelevant information, and the focal loss that can highly mitigate the class imbalance issue. 
\section{Conclusions}
\label{sec:length}
In this paper, we propose a novel non-temporal mechanism to enhance the ABSA task through modeling inter-aspect dependencies, which can selectively incorporate neighboring aspects related information into target-specific representation. 
Furthermore, we introduce focal loss to address the class imbalance issue for the first time. 
Extensive experiments have demonstrated the effectiveness of our proposed approach on both restaurant and laptop domains.

In the future, we would like to explore the effectiveness of our approach in other tasks~\cite{liang2020infusing,liang-etal-2021-modeling,liang-etal-2021-towards}.


\section*{Acknowledgements}
Liang, Chen and Xu are supported by the National
Natural Science Foundation of China (Contract
61370130, 61976015, 61976016 and 61876198),
and the Beijing Municipal Natural Science Foundation (Contract 4172047).

\bibliography{emnlp-ijcnlp-2019}
\bibliographystyle{acl_natbib}

\end{document}